\documentclass[runningheads]{llncs}
\usepackage{marvosym}
\usepackage{makeidx}  
\usepackage{url}
\usepackage{color}
\usepackage{longtable}
\usepackage{multirow}
\usepackage{graphicx}
\usepackage{amssymb, amsmath, bm}
\usepackage{mathtools}
\usepackage{mathrsfs}
\usepackage{cite}
\usepackage{dsfont}
\usepackage[colorlinks,linkcolor=blue]{hyperref}
\usepackage{makecell}
\usepackage[noend]{algpseudocode}

\usepackage{algorithmicx,algorithm}
\usepackage{graphicx}

\usepackage{multirow}
\usepackage{diagbox}
\begin{document}
\title{Agent with Tangent-based Formulation and Anatomical Perception for Standard Plane Localization in 3D Ultrasound}
\titlerunning{Agent for Standard Plane Localization in 3D Ultrasound}
\author{Yuxin Zou\inst{1,2,3}\thanks{Yuxin Zou and Haoran Dou contribute equally to this work.} \and Haoran Dou\inst{4,5\star} \and Yuhao Huang \inst{1,2,3} \and Xin Yang \inst{1,2,3} \and Jikuan Qian\inst{6} \and Chaojiong Zhen \inst{7} \and Xiaodan Ji \inst{7} \and Nishant Ravikumar~\inst{4,5} \and Guoqiang Chen \inst{7} \and Weijun Huang \inst{7} \and Alejandro F. Frangi\inst{4,5,8,9} \and Dong Ni\inst{1,2,3}\textsuperscript{(\Letter)}}
%
\authorrunning{Y. Zou et al.}
%
\institute{
National-Regional Key Technology Engineering Laboratory for Medical Ultrasound, School of Biomedical Engineering, Health Science Center, Shenzhen University, China \\
\email{nidong@szu.edu.cn}
\and Medical Ultrasound Image Computing (MUSIC) Lab, Shenzhen University, China \and
Marshall Laboratory of Biomedical Engineering, Shenzhen University, China \and
Centre for Computational Imaging and Simulation Technologies in Biomedicine (CISTIB), School of Computing, University of Leeds, Leeds, UK \and
Biomedical Imaging Department, Leeds Institute for Cardiovascular and Metabolic Medicine (LICAMM), School of Medicine, University of Leeds, Leeds, UK
\and
Shenzhen RayShape Medical Technology Co., Ltd, China
\and
Department of Ultrasound, The First People's Hospital of Foshan, Foshan, China
\and
Departments of Cardiovascular Sciences and Electrical Engineering, KU Leuven, Leuven, Belgium
\and
Alan Turing Institute, London, UK
}%
\maketitle   
\begin{abstract}
Standard plane (SP) localization is essential in routine clinical ultrasound (US) diagnosis. Compared to 2D US, 3D US can acquire multiple view planes in one scan and provide complete anatomy with the addition of coronal plane. However, manually navigating SPs in 3D US is laborious and biased due to the orientation variability and huge search space. In this study, we introduce a novel reinforcement learning (RL) framework for automatic SP localization in 3D US. Our contribution is three-fold.  First, we formulate SP localization in 3D US as a tangent-point-based problem in RL to restructure the action space and significantly reduce the search space. Second, we design an auxiliary task learning strategy to enhance the model’s ability to recognize subtle differences crossing Non-SPs and SPs in plane search. Finally, we propose a spatial-anatomical reward to effectively guide learning trajectories by exploiting spatial and anatomical information simultaneously. We explore the efficacy of our approach on localizing four SPs on uterus and fetal brain datasets. The experiments indicate that our approach achieves a high localization accuracy as well as robust performance.
\end{abstract}

\section{Introduction}
Ultrasound (US) is the primary scanning method in routine diagnosis due to its lack of radiation, real-time imaging, low cost and high mobility~\cite{beyer2020scans}. As the preliminary step in US diagnosis, acquiring standard plane (SP) provides anatomical content for subsequent bio-marker measurement and abnormal diagnosis. Compared with the 2D US, 3D US shows natural superiority in acquiring multiple view planes via a single scan and providing complete and fruitful 3D information~\cite{dou2019agent}. Furthermore, 3D US enables the sonographer to obtain additional SP that is unavailable using the 2D US owing to bony pelvis~\cite{turkgeldi2015role}, e.g., the coronal plane of the uterus, which is important for assessing uterine abnormalities. However, manually localizing SP is laborious and biased due to the huge search space and orientation variability of 3D US. The internal invisibility of US volume makes it further difficult for sonographers to search the vast 3D space. Hence, developing an automatic approach for localizing SP in 3D US is highly desirable to relieve the burden of sonographers and reduce operator dependency.

Regression of plane parameters or transformation matrix is a common strategy for 3D SP localization~\cite{chykeyuk2013class, li2018standard, lorenz2018automated, yeung2021learning}. Chykeyuk~\textit{et al.}~\cite{chykeyuk2013class} proposed a random forest method to regress plane parameters in 3D echocardiography. Li~\textit{et al.}~\cite{li2018standard} introduced a deep neural network to localize fetal brain planes by computing transformation matrix iteratively. Lorenz~\textit{et al.}~\cite{lorenz2018automated} proposed to extract the abdomen plane through anatomical landmark detection and align them to a fetal organ model. Most recently, Yeung~\textit{et al.}~\cite{yeung2021learning} designed an annotation-efficient approach to learn the mapping between the 2D images and 3D space in US.
However, all aforementioned methods are limited by the difficulty in optimizing a highly abstract mapping function~\cite{yang2021searching}.

Recently, reinforcement learning (RL) shows great potential in addressing the SP localization problem by its specific reward mechanism and interactive planning~\cite{alansary2018automatic}. Dou~\textit{et al.}~\cite{dou2019agent, yang2021agent} first proposed the RL-based framework for localizing SP in 3D US. They designed a registration-based warm-up strategy to address the large orientation variability of US volume and provide effective initialization for the agent in RL. In their follow-up work~\cite{yang2021searching}, they embedded neural network searching in the RL optimization and designed a multi-agent collaborative system for multi-SP navigation. Motivated by~\cite{dou2019agent, yang2021agent}, several works~\cite{li2021image, li2021autonomous} employed the RL to achieve autonomous navigation of US probe towards SP. Although these approaches achieved high performance in SP localization, several issues are still required to be addressed. First, current studies~\cite{dou2019agent,yang2021agent,yang2021searching} rely on initial registration to ensure data orientation consistency. They are easily trapped when pre-registration fails. Second, most studies~\cite{dou2019agent,yang2021agent,yang2021searching,alansary2018automatic} designed an eight-dimensional action space in terms of angle and distance, where the coupling among the directional cosines in the formula and the huge search space make the optimization difficult. Third, existing RL systems~\cite{dou2019agent,yang2021agent,yang2021searching,alansary2018automatic} are only driven by the plane-movement-based reward function, lacking the perception and guidance of anatomical structures.

To address the outstanding issues mentioned above, we introduce a novel RL-based framework for automatic SP localization in 3D US. In particular, we define a new tangent-point-based plane formulation in RL to restructure the action space and significantly reduce the search space; we design an auxiliary task learning strategy to enhance the model's ability to recognize subtle differences crossing Non-SPs and SPs in plane search; we propose a spatial-anatomical reward to effectively guide learning trajectories by exploiting spatial and anatomical information simultaneously.

\section{Method}
\begin{figure*}[!htp]
	\centering
	\includegraphics[width=0.88\linewidth]{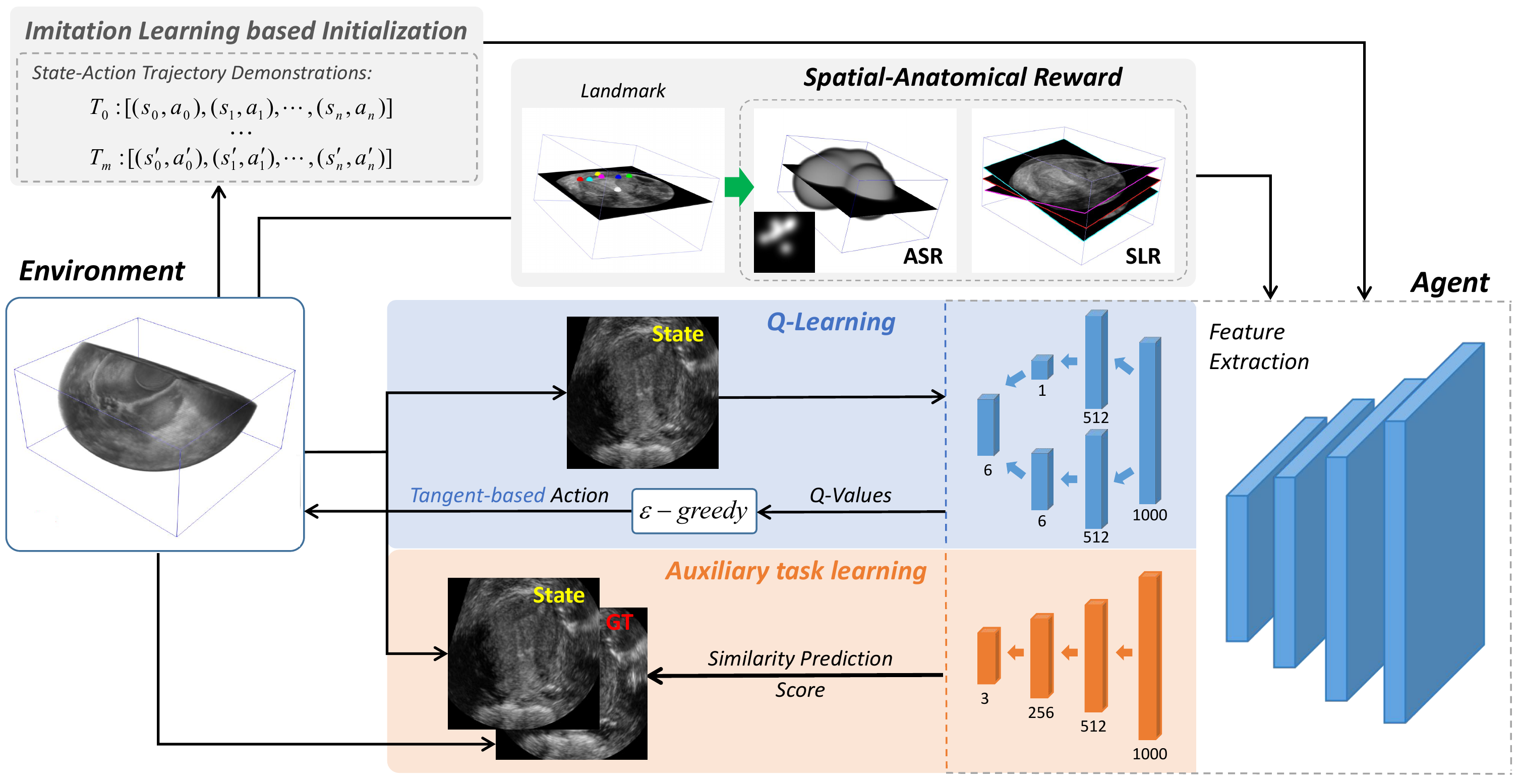}
	\caption{Overview of the proposed SP localization framework. SLR: spatial location reward; ASR: anatomical structure
reward.}
	\label{fig:framework}
\end{figure*}

As shown in Fig.~\ref{fig:framework}, our proposed SP localization framework is based on RL, where the \textit{agent} (neural network) interacts with the \textit{environment} (3D US Volume) to learn an optimal SP searching policy with the maximum accumulated \textit{reward}. Additionally, we equip the RL framework with an auxiliary task to predict the similarity of the current state and target state, thus boosting the model's recognition ability. An imitation learning module is leveraged to initialize the agent in RL framework for speeding up the optimization.
\subsection{Reinforcement Learning for Plane Localization}
\begin{figure*}[t]
	\centering
	\includegraphics[width=0.88\linewidth]{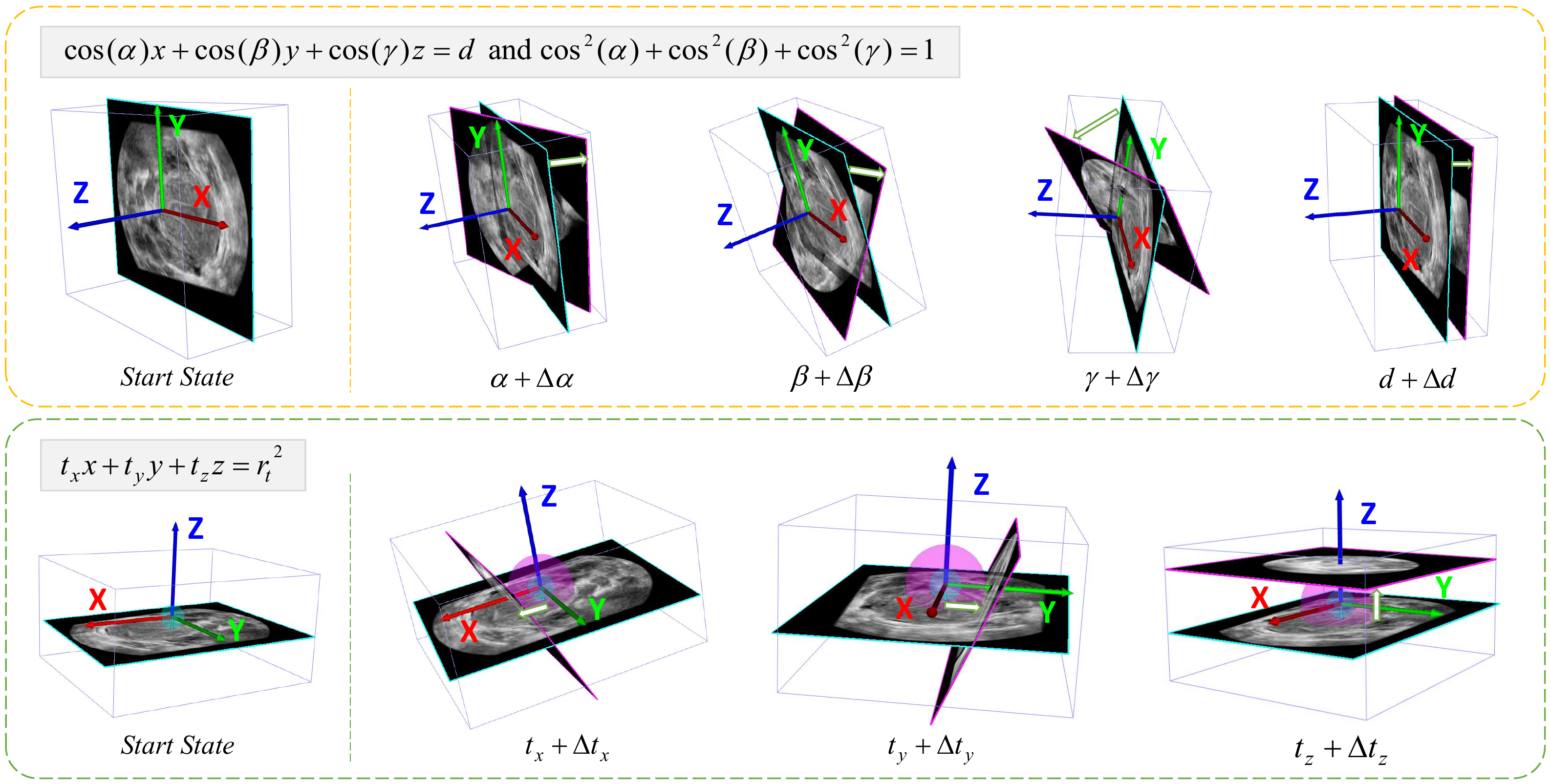}
	\caption{Comparison of our formulation (bottom) and previous one (top). Previous formulation controls the plane movement by adjusting directional cosines ($\alpha, \beta, \gamma$) and the distance to the origin ($d$). Instead, our design modifies the plane movement by translating the coordinate ($t_x, t_y, t_z$) of the tangent point.}
	\label{fig:formulation}
\end{figure*}

Formulation of the SP localization is essential for optimizing the RL framework. Previous works~\cite{alansary2018automatic, dou2019agent, yang2021agent, yang2021searching} modeled the plane movement by adjusting the plane function in terms of normal and distance (see Fig.~\ref{fig:formulation}). However, the coupling among the directional cosines ($cos^2(\alpha) + cos^2(\beta) + cos^2(\gamma) = 1$) makes actions dependent and unable to reflect the model's objective accurately, resulting in obstacles to agent learning. Furthermore, the RL training can easily fail without the pre-registration processing~\cite{dou2019agent} to limit orientation variability and search space~\cite{yang2021agent}. This study proposes a novel tangent-point-based formulation for SP localization in RL. Our formulation builds a simplified and mutual-independent action space to improve the optimization of the RL framework, enabling accurate SP localization even within the unaligned US environment. Fig.~\ref{fig:formulation} illustrates the comparison of our formulation and that of previous works. We discovered that any view plane in the 3D space can be defined uniquely as its tangent point ($t_x$, $t_y$, $t_z$) on the sphere centering in the origin with the radius of $r_t$, where $r_t^2=t_x^2 + t_y^2 + t_z^2$. The plane function can be written as $t_x x+t_y y + t_z z = r_t^2$. Therefore, SP localization can be re-formulated into the tangent point searching task, where the action space only contains the translation of the coordinate of the tangent point. The proposed formulation is unrestricted by directional cosines coupling with less action space than the previous ones (6 $<$ 8), facilitating agent learning. We explain the details of the elements in the RL here.

\subsubsection{Action}
The action space is defined as $\{\pm a_{t_x},\pm a_{t_y},\pm a_{t_z}\}$ based on our formulation. Given an action in step $i$, the tangent point coordinate can be modified accordingly, e.g. $t_x^{i+1}=t_x^{i}+a_{t_x}$. We noticed experimentally that the image content is sensitive to the step size when the corresponding sphere radius is small. To address this issue, we model the agent-environment interaction as a multi-stage motion process by progressively scaling down the step size from 1.0 to 0.01 when the agent appears to oscillate for three steps. We terminate the agent searching at 60 steps.

\subsubsection{State}
The state is defined as the 2D US image of size $224^2$ reconstructed by sampling the voxels from the volume according to the plane parameters. We concatenate the two images obtained from the previous two iterations with the current image to enrich the state information~\cite{mnih2015human}.

\subsubsection{Rewards}
The reward function instructs the agent on the optimal searching policy with the proper action. Recent works~\cite{yang2021agent, yang2021searching} calculated the reward function based on the differences of parameters in the defined plane function between adjacent iterations. Although effective, we argue that such design may cause the agent to lack anatomical perception and guidance, which may affect the agent's performance on the abnormal data. In this study, we design a spatial-anatomical reward, involving 1) spatial location reward (SLR) and 2) anatomical structure reward (ASR). Specifically, SLR motivates the agent to approach the target location by minimizing the Euclidean distance of the plane parameters between the current plane and target plane, while ASR encourages the agent to perceive anatomical information. We construct the heatmap with a Gaussian kernel at anatomical landmarks (see Fig.~\ref{fig:framework}) to calculate the ASR. The reward can be defined as:
\begin{equation}
	r= sgn({\parallel P_{t-1}-P_{g}\parallel}_{2}-{\parallel P_{t}-P_{g}\parallel}_{2})+sgn({\mid I_{t}-I_{g}\mid} - \mid I_{t-1}-I_{g}\mid)
	\label{eq:reward}
\end{equation}
where $sgn(\cdot)$ is the sign function, $P_t$ and $P_g$ indicate the tangent point parameters of the prediction plane and the target plane, respectively. Likewise, $I_t$ and $I_g$ represent the sum of the heatmap value corresponding to the prediction plane and the target plane, respectively.

\subsubsection{Loss function}
Similar to~\cite{yang2021agent, yang2021searching}, we perform the dueling Q-learning~\cite{wang2016dueling} to train the agent. Given the prioritized replay buffer $\mathcal{M}$, which stores the transitions of each step, including state $s$, action $a$, reward $r$, and next state $s'$, the loss function for the Q-learning part of our framework can be defined as:
\begin{small}
\begin{equation}
	\mathcal{L}_{Q}(\omega) = \mathbb{E}_{s,r,a,s' \sim U(\mathcal{M})}(r + \gamma Q_{target}(s',\mathop {argmax}\limits_{a'}Q(s',a';\omega);\omega')-Q(s,a;\omega))^{2}
	\label{eq:loss_q}
\end{equation}
\end{small}
where $\gamma$ is the discount factor that balances the importance of current and future rewards. $s$, $s'$ $a$ and $a'$ are the state and action in the current/next step. In this study, the current and target Q-network share the same network architecture (i.e. ResNet~\cite{he2016deep}), and $w$ and $w'$ are their parameters. During training, the target Q-network copied the parameters of the current Q-network every 1800 steps.

\subsection{Auxiliary Task of State-Content Similarity Prediction}
Localizing SPs in 3D US is challenging due to the low inter-class variability between SPs and non-SPs in the searching procedure and high intra-class variability of SPs. Most approaches\cite{li2018standard,yang2021agent,yang2021searching} lack the proper strategy to use the image-level content information (e.g., anatomical priors), resulting in inefficient data utilization and agent learning. Auxiliary tasks for RL\cite{jaderberg2016reinforcement, mirowski2016learning} could improve learning efficiency and boost the performance by learning the fine-grained representations. To facilitate the agent to learn the content representations, we design an auxiliary task of state-content similarity prediction (SCSP). As shown in Fig.~\ref{fig:framework}, we utilize an additional regression branch in the agent network to predict the similarity of the current state to the target state. The content similarity is measured by normalized cross-correlation (NCC)~\cite{yoo2009fast}. The loss function for the auxiliary task part of our framework is defined as:
\begin{equation}
	\mathcal{L}_{A}(\omega) = \mathbb{E}_{s \sim U(\mathcal{M})} \parallel Score_{gt}(s, s_{gt})-Score_{pre}(s;\omega) \parallel_{2}
	\label{eq:loss_a}
\end{equation}
where $Score_{gt}(s, s_{gt})$ is the NCC between the current state $s$ and the target state $s_{gt}$; $Score_{pre}(s; \omega)$ is the NCC score predicted by SCSP. Overall, the total loss function of our proposed RL framework is $\mathcal{L} = \mathcal{L}_{Q}+\delta \mathcal{L}_{A}$, where $\delta=0.5$ is the weight to balance the importance of Q-learning loss and auxiliary task loss.

\subsection{Imitation Learning based Initialization}
It is difficult for the agent to obtain effective samples during interacting with the unaligned US environment because the replay buffer will store a number of futile data during the agent exploring the 3D space. It might reduce the agent learning efficiency even harm the performance. \cite{hester2018deep} pointed out that imitation learning could effectively address this issue by pre-training the agent to ensure it could gain enough knowledge before exploring the environment, thus boosting the learning efficiency. Therefore, we adopt imitation learning as an initialization of the agent. Specifically, we first randomly select 20 initial tangent points in each training data and then approach the target plane by taking the optimal action in terms of the distance to the target tangent point. This can imitate the expert's operation and obtain a number of effective demonstrated state-action trajectories (e.g., {$(s_0,a_0), (s_1,a_1),\dots,(s_n, a_n)$}). After that, we perform the supervised learning with the cross-entropy loss on the agent based on the randomly sampled state-action pair. With a well-trained agent based on imitation learning initialization, the learning of the RL framework can be eased and accelerated.

\section{Experimental Result}
\subsection{Materials and Implementation Details}
We validated our proposed method on four SPs in two datasets, including the coronal (C) plane in the uterus and the trans-ventricular (TV), trans-thalamic (TT), and trans-cerebellar (TC) plane in the fetal brain. The uterus dataset has 363 normal patients and 45 abnormal patients (Congenital Uterine Anomalies, CUAs) with an average volume size of $432\times377\times217$ and spacing of $0.3\times0.3\times0.3mm^3$; the fetal brain dataset has 432 patients with an average volume size of $270\times207\times235$ and spacing of $0.5\times0.5\times0.5mm^3$. Six experienced sonographers manually annotated SPs and landmarks using the Pair annotation software package~\cite{liang2022sketch} under strict quality control. We randomly split each dataset for training, validating, testing of 290, 20, 53 in the uterus, and 330, 30, 72 in the fetal brain, respectively. To verify the generalizability of our method, we only involve the healthy subjects in our training dataset and test the 45 CUAs patients independently.

In this study, we implemented our method by PyTorch using a standard PC with an NVIDIA RTX 2080Ti GPU. We trained the model through Adam optimizer with a learning rate of 5e-5 and a batch size of 32 for 100 epochs. The discount factor $\gamma$ in Eq.~\ref{eq:loss_q} was set as 0.85. The size of the prioritized Replay Buffer was set as 15000. The $\epsilon -greedy$ exploration strategy was set according to~\cite{yang2021agent}. We calculated the mean ($\mu$) and standard deviation ($\sigma$) of target tangent point locations in the training dataset and randomly initialized start points for training within $\mu \pm 2\sigma$ to capture 95$\%$ variability approximately. For testing, the origin was set as the initial tangent point.

\begin{table}[t] 
\renewcommand\arraystretch{0.9}
    \caption{Quantitative comparison of different methods on SP localization. $C_N$ and $C_P$ mean the coronal plane of normal and abnormal uteruses, respectively. (mean$\pm$std)}\label{tab:compare_different}
	\centering
	\begin{tabular}{cc|c|c|c|c|c|c}
		\hline
        \hline
		\multicolumn{2}{c|}{\diagbox{Metrics}{Methods}} &
		\multicolumn{1}{c|}{{$RG_\mathit{single}$}} & \multicolumn{1}{c|}{{$RG_\mathit{ITN}$}} & \multicolumn{1}{c|}{{$Regist$}} & \multicolumn{1}{c|}{{$RL_\mathit{AVP}$}} & \multicolumn{1}{c|}{{$RL_\mathit{WSADT}$}} & \multicolumn{1}{c}{{$RL_\mathit{Ours}$}} \\
		\hline
		
		\multicolumn{1}{c|}{\multirow{4}{*}{\quad \scriptsize{$C_{N}$} \quad}} &
		\scriptsize{Ang(°)} &
		    \scriptsize{{16.24$\pm$9.72}} &
		    \scriptsize{{29.61$\pm$24.49}} &
            \scriptsize{{15.77$\pm$14.34}} &
            \scriptsize{{49.15$\pm$13.26}} &
            \scriptsize{{15.71$\pm$14.99}} &
            \scriptsize{\textbf{10.36$\pm$11.92}} \\

		\multicolumn{1}{c|}{} &
		\scriptsize{Dis(mm)} &
            \scriptsize{{2.48$\pm$2.07}} &
            \scriptsize{\textbf{0.83$\pm$0.75}} &
            \scriptsize{{1.94$\pm$1.90}} &
            \scriptsize{{2.86$\pm$2.10}} &
            \scriptsize{{1.83$\pm$1.91}} &
            \scriptsize{{0.88$\pm$0.84}} \\

		\multicolumn{1}{c|}{} &
		\scriptsize{SSIM} &
            \scriptsize{{0.10$\pm$0.07}} &
            \scriptsize{{0.55$\pm$0.11}} &
            \scriptsize{{0.46$\pm$0.10}} &
            \scriptsize{{0.28$\pm$0.06}} &
            \scriptsize{{0.48$\pm$0.12}} &
            \scriptsize{\textbf{0.61$\pm$0.19}} \\

        \multicolumn{1}{c|}{} &
		\scriptsize{NCC} &
            \scriptsize{{0.51$\pm$0.17}} &
            \scriptsize{{0.64$\pm$0.16}} &
            \scriptsize{{0.69$\pm$0.14}} &
            \scriptsize{{0.36$\pm$0.16}} &
            \scriptsize{{0.70$\pm$0.16}} &
            \scriptsize{\textbf{0.74$\pm$0.18}} \\
        \hline

		\multicolumn{1}{c|}{\multirow{4}{*}{\quad \scriptsize{$C_{P}$} \quad}} &
		\scriptsize{Ang(°)} &
		\scriptsize{{16.25$\pm$8.16}} &
        \scriptsize{{29.98$\pm$25.67}} &
        \scriptsize{{21.79$\pm$19.30}} &
        \scriptsize{{52.58$\pm$13.07}} &
        \scriptsize{{21.22$\pm$19.20}} &
        \scriptsize{\textbf{11.90$\pm$5.78}} \\
		
		\multicolumn{1}{c|}{} &
		\scriptsize{Dis(mm)} &
		\scriptsize{{2.88$\pm$2.37}} &
        \scriptsize{{1.50$\pm$1.26}} &
        \scriptsize{{2.64$\pm$2.57}} &
        \scriptsize{{4.30$\pm$2.67}} &
        \scriptsize{{2.76$\pm$2.53}} &
        \scriptsize{\textbf{1.48$\pm$1.35}} \\

		\multicolumn{1}{c|}{} &
		\scriptsize{SSIM} &
		\scriptsize{{0.09$\pm$0.08}} &
        \scriptsize{{0.51$\pm$0.12}} &
        \scriptsize{{0.41$\pm$0.10}} &
        \scriptsize{{0.27$\pm$0.06}} &
        \scriptsize{{0.41$\pm$0.11}} &
        \scriptsize{\textbf{0.52$\pm$0.14}} \\
		
		\multicolumn{1}{c|}{} &
		\scriptsize{NCC} &
		\scriptsize{{0.49$\pm$0.18}} &
        \scriptsize{{0.58$\pm$0.17}} &
        \scriptsize{{0.61$\pm$0.17}} &
        \scriptsize{{0.30$\pm$0.19}} &
        \scriptsize{{0.61$\pm$0.17}} &
        \scriptsize{\textbf{0.62$\pm$0.20}} \\
		
		\hline
		
		\multicolumn{1}{c|}{\multirow{4}{*}{\quad \scriptsize{TT} \quad}} &
		\scriptsize{Ang(°)} &
		\scriptsize{{30.46$\pm$21.05}} &
        \scriptsize{{21.15$\pm$20.46}} &
        \scriptsize{{14.37$\pm$13.42}} &
        \scriptsize{{54.05$\pm$15.35}} &
        \scriptsize{\textbf{10.48$\pm$5.80}} &
        \scriptsize{{10.89$\pm$7.70}} \\

		\multicolumn{1}{c|}{} &
		\scriptsize{Dis(mm)} &
		\scriptsize{{3.53$\pm$2.19}} &
        \scriptsize{{0.94$\pm$0.75}} &
        \scriptsize{{2.12$\pm$1.42}} &
        \scriptsize{{4.34$\pm$2.97}} &
        \scriptsize{{2.02$\pm$1.33}} &
        \scriptsize{\textbf{0.80$\pm$0.93}} \\

		\multicolumn{1}{c|}{} &
		\scriptsize{SSIM} &
		\scriptsize{{0.58$\pm$0.14}} &
        \scriptsize{{0.85$\pm$0.05}} &
        \scriptsize{{0.83$\pm$0.09}} &
        \scriptsize{{0.64$\pm$0.06}} &
        \scriptsize{{0.78$\pm$0.06}} &
        \scriptsize{\textbf{0.92$\pm$0.06}} \\

		\multicolumn{1}{c|}{} &
		\scriptsize{NCC} &
		\scriptsize{{0.51$\pm$0.27}} &
        \scriptsize{{0.57$\pm$0.17}} &
        \scriptsize{\textbf{0.83$\pm$0.13}} &
        \scriptsize{{0.44$\pm$0.10}} &
        \scriptsize{{0.78$\pm$0.14}} &
        \scriptsize{{0.78$\pm$0.22}} \\
		\hline

		\multicolumn{1}{c|}{\multirow{4}{*}{\quad \scriptsize{TV} \quad}} &
		\scriptsize{Ang(°)} &
		\scriptsize{{38.78$\pm$25.40}} &
        \scriptsize{{26.80$\pm$25.55}} &
        \scriptsize{{13.40$\pm$4.68}} &
        \scriptsize{{53.77$\pm$14.58}} &
        \scriptsize{{10.39$\pm$4.03}} &
        \scriptsize{\textbf{8.65$\pm$7.10}} \\

		\multicolumn{1}{c|}{} &
		\scriptsize{Dis(mm)} &
		\scriptsize{{7.44$\pm$5.99}} &
        \scriptsize{{1.43$\pm$0.98}} &
        \scriptsize{{2.68$\pm$1.58}} &
        \scriptsize{{4.27$\pm$2.73}} &
        \scriptsize{{2.48$\pm$1.27}} &
        \scriptsize{\textbf{1.16$\pm$2.45}} \\

		\multicolumn{1}{c|}{} &
		\scriptsize{SSIM} &
		\scriptsize{{0.58$\pm$0.11}} &
        \scriptsize{{0.85$\pm$0.05}} &
        \scriptsize{{0.71$\pm$0.11}} &
        \scriptsize{{0.64$\pm$0.06}} &
        \scriptsize{{0.66$\pm$0.14}} &
        \scriptsize{\textbf{0.92$\pm$0.06}} \\

		\multicolumn{1}{c|}{} &
		\scriptsize{NCC} &
        \scriptsize{{0.53$\pm$0.18}} &
        \scriptsize{{0.56$\pm$0.16}} &
        \scriptsize{{0.56$\pm$0.27}} &
        \scriptsize{{0.43$\pm$0.11}} &
        \scriptsize{{0.57$\pm$0.28}} &
        \scriptsize{\textbf{0.78$\pm$0.25}} \\
		\hline

		\multicolumn{1}{c|}{\multirow{4}{*}{\quad \scriptsize{TC} \quad}} &
		\scriptsize{Ang(°)} &
		\scriptsize{{33.08$\pm$21.18}} &
        \scriptsize{{27.21$\pm$23.83}} &
        \scriptsize{{16.24$\pm$13.57}} &
        \scriptsize{{52.70$\pm$16.03}} &
        \scriptsize{{10.26$\pm$7.25}} &
        \scriptsize{\textbf{9.75$\pm$8.45}} \\

		\multicolumn{1}{c|}{} &
		\scriptsize{Dis(mm)} &
		\scriptsize{{3.82$\pm$3.30}} &
        \scriptsize{{1.26$\pm$1.06}} &
        \scriptsize{{3.47$\pm$2.39}} &
        \scriptsize{{4.20$\pm$2.65}} &
        \scriptsize{{2.52$\pm$2.13}} &
        \scriptsize{\textbf{0.88$\pm$1.15}} \\

		\multicolumn{1}{c|}{} &
		\scriptsize{SSIM} &
        \scriptsize{{0.59$\pm$0.13}} &
        \scriptsize{{0.84$\pm$0.05}} &
        \scriptsize{{0.68$\pm$0.18}} &
        \scriptsize{{0.64$\pm$0.07}} &
        \scriptsize{{0.64$\pm$0.14}} &
        \scriptsize{\textbf{0.88$\pm$0.09}} \\

		\multicolumn{1}{c|}{} &
		\scriptsize{NCC} &
		\scriptsize{{0.56$\pm$0.22}} &
        \scriptsize{{0.55$\pm$0.16}} &
        \scriptsize{{0.55$\pm$0.29}} &
        \scriptsize{{0.45$\pm$0.12}} &
        \scriptsize{{0.55$\pm$0.30}} &
        \scriptsize{\textbf{0.69$\pm$0.25}} \\
		
		\hline
		\hline
	\end{tabular}
\end{table}

\begin{table}[t]
	\caption{Ablation study for analyzing SCSP and ASR.}\label{tab:ablation}
	\centering
	\resizebox{\textwidth}{14mm}{
    	\begin{tabular}{c|c|c|c|c|c|c|c|c|c}
    		\hline
    		\hline
    		\multicolumn{2}{c|}{{Strategy}} & \multicolumn{4}{c|}{{$C_{N}$}} & \multicolumn{4}{c}{{$C_{P}$}}\\
    		\cline{1-10} \multicolumn{1}{m{0.7cm}|}{\centering \scriptsize{SCSP}} & \multicolumn{1}{m{0.6cm}|}{\centering \scriptsize{ASR}} & \scriptsize{\textit{Ang(◦)$\downarrow$}} & \scriptsize{\textit{ Dis(mm)$\downarrow$}} & \scriptsize{\textit{SSIM$\uparrow$}} & \scriptsize{\textit{NCC$\uparrow$}} & \scriptsize{\textit{Ang(◦)$\downarrow$}} & \scriptsize{\textit{ Dis(mm)$\downarrow$}} & \scriptsize{\textit{SSIM$\uparrow$}} &  \scriptsize{\textit{NCC$\uparrow$}} \\
    		\hline
    		$\times$ & $\times$ & \scriptsize{{14.11$\pm$11.08}} & \scriptsize{{0.91$\pm$0.87}} & \scriptsize{{0.57$\pm$0.17}} & \scriptsize{{0.69$\pm$0.17}} & \scriptsize{{15.57$\pm$10.95}} & \scriptsize{{1.45$\pm$1.25}} & \scriptsize{{0.51$\pm$0.14}} &\scriptsize{{0.59$\pm$0.17}} \\
    		\hline
    		$\surd$ & $\times$ & \scriptsize{{11.83$\pm$13.54}} & \scriptsize{{0.84$\pm$0.74}} & \scriptsize{{0.58$\pm$0.18}} & \scriptsize{{0.70$\pm$0.19}} &
    		\scriptsize{{13.80$\pm$9.13}} & \scriptsize{{1.63$\pm$1.61}} &
    		\scriptsize{{0.52$\pm$0.14}} &\scriptsize{{0.59$\pm$0.22}} \\
    		\hline
    		$\times$ & $\surd$ & \scriptsize{{12.72$\pm$10.78}} & \scriptsize{\textbf{{0.80$\pm$0.72}}} & \scriptsize{{0.59$\pm$0.16}} & \scriptsize{{0.70$\pm$0.17}} &
    		\scriptsize{{14.09$\pm$7.89}} & \scriptsize{\textbf{1.27$\pm$1.07}} & \scriptsize{{0.51$\pm$0.14}} &\scriptsize{{0.59$\pm$0.20}}  \\
    		\hline
    		$\surd$ & $\surd$ & \scriptsize{\textbf{10.36$\pm$11.92}} & \scriptsize{{0.88$\pm$0.84}} & \scriptsize{\textbf{0.61$\pm$0.19}} & \scriptsize{\textbf{0.74$\pm$0.18}} & \scriptsize{\textbf{11.90$\pm$5.78}} & \scriptsize{{1.48$\pm$1.35}} & \scriptsize{\textbf{0.52$\pm$0.13}} & \scriptsize{\textbf{0.62$\pm$0.20}} \\
    		\hline
    		\hline
    	\end{tabular}}
\end{table}

\begin{figure*}[t]
	\centering
	\includegraphics[width=0.9\linewidth]{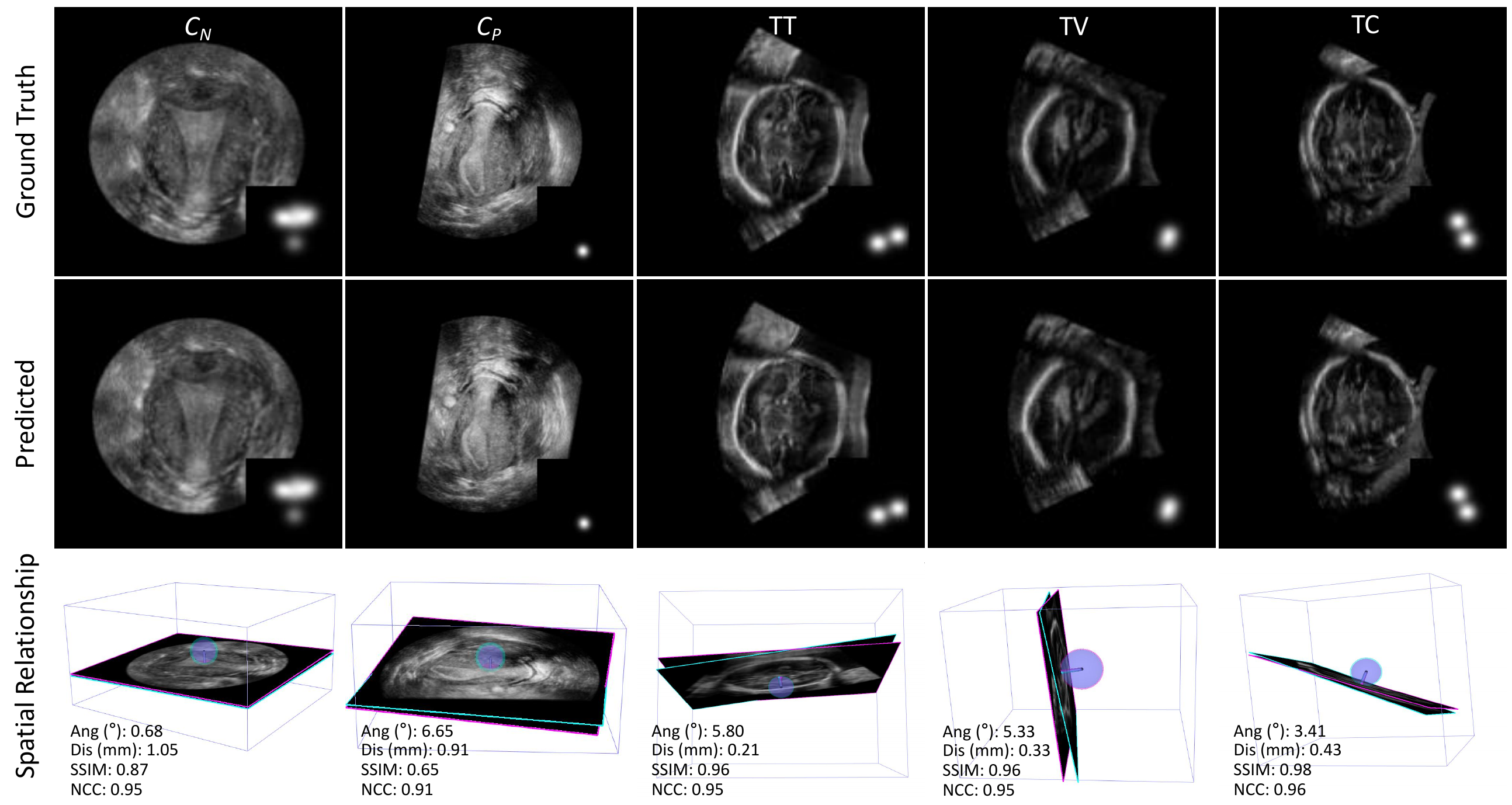}
    \caption{Visual example results of our methods. The first row shows the ground truth of five SPs; the second row shows the predicted plane with its landmark heatmap in the Lower right corner; the third row shows the 3D spatial relationship between ground truth (red) and prediction (green).}
	\label{fig:result}
\end{figure*}

\subsection{Quantitative and Qualitative Analysis}
To demonstrate the efficacy of our proposed method, we performed the comparison with five SOTA approaches including regression-based, (i.e., $RG_{Single}$, $RG_\mathit{ITN}$~\cite{li2018standard}), registration-based (i.e., $Regist$~\cite{dou2019agent}), and RL-based methods (i.e., $RL_\mathit{AVP}$~\cite{alansary2018automatic}, $RL_\mathit{WSADT}$~\cite{yang2021agent}). Four criteria, including the spatial metrics (angle and distance between two planes, Ang $\&$ Dis) and content metrics (Structural Similarity Index and Normalized Cross-correlation, SSIM $\&$ NCC), were used to evaluate the performance.

As shown in Table~\ref{tab:compare_different}, our proposed method outperforms all of the others on most of the metrics, indicating the superior ability of our method in SP localization tasks. Specifically, we can observe that RL gains large boosting in performance through pre-registration to ensure orientation consistency ($RL_\mathit{AVP}$ vs. $RL_\mathit{WSADT}$). In comparison, our new formulation could enable the RL algorithm to achieve superiority even without pre-registration. Additionally, prior RL-based methods fail easily in the abnormal dataset ($RL_\mathit{WSADT}$ in $C_P$). On the contrary, our method obtains consistent performance on both normal and abnormal datasets. Table~\ref{tab:ablation} shows the results of the ablation study to investigate the impact of each designed module. It can be observed that SCSP improves the generalizability of the model by enhancing the recognition of the SPs and non-SPs. It is beneficial to compose ASR with the basic reward SLR to boost agents' perception of anatomical structures, which enables our model to generalize external abnormal uterus dataset having significant content differences with the normal one.  Visual illustration of the results of our method in Fig.~\ref{fig:result} also shows the extent of the SP localization performance associated with the quantitative measures reported in Table~\ref{tab:compare_different}.

\section{Conclusion}
We proposed a novel RL framework for SP localization in 3D US. We define a tangent-point-based plane formulation to restructure action space and improve agent optimization within unaligned US environment. This formulation can be extended to similar tasks in other modalities, e.g., CT or MRI. We propose a content-aware regression auxiliary task to improve the agent's robustness to noisy US environment. In addition, we design a spatial-anatomical reward to provide both spatial and anatomical knowledge for the agent. Moreover, we initialize the agent by imitation learning to improve the training efficiency. Experiments show that our method can achieve superior performance for localizing four SPs in two unaligned datasets including abnormal cases, which indicates its great potential for localizing SPs in randomly initialized spaces and abnormal cases.
\subsubsection{Acknowledgement.}
This work was supported by the grant from National Natural Science Foundation of China (Nos. 62171290, 62101343), Shenzhen-Hong Kong Joint Research Program (No. SGDX20201103095613036), Shenzhen Science and Technology Innovations Committee (No. 20200812143441001), the Royal Academy of Engineering (INSILEX CiET1819/19), the Royal Society Exchange Programme CROSSLINK IES$\backslash$NSFC$\backslash$201380, and Engineering and Physical Sciences Research Council (EPSRC) programs TUSCA EP/V04799X/1.

%
%
%
\bibliographystyle{splncs04}
\bibliography{paper1248}

\end{document}